\documentclass[a4, 10pt, conference]{ieeeconf}      

\IEEEoverridecommandlockouts                              
                                                          
\usepackage{amsmath,amssymb,mathtools} 
\usepackage{dsfont,eucal,bbm,bm,nicefrac} 
\usepackage{graphicx,float} 
	\graphicspath{{./images/}}
\usepackage{soul}
\usepackage{float}
\usepackage{caption}
\usepackage{subcaption}
\usepackage[ruled,vlined]{algorithm2e}
\usepackage{cases}

\usepackage{tikz}
\usepackage{textcomp}
\usepackage{hyperref}
\usepackage{lipsum}

\newtheorem{remark}{Remark}

\newcommand{\T}{\ensuremath{\mathrm{T}}} 

\def\BibTeX{{\rm B\kern-.05em{\sc i\kern-.025em b}\kern-.08em
    T\kern-.1667em\lower.7ex\hbox{E}\kern-.125emX}}

\title{\LARGE \bf Cooperative Bidirectional Mixed-Traffic Overtaking
}

\author{Faizan M. Tariq$^{1*}$, Nilesh Suriyarachchi$^{1*}$, Christos Mavridis$^{1}$ and John S. Baras$^{1}$
\thanks{$^{1}$Electrical and Computer Engineering Department and the Institute for Systems Research, University of Maryland, College Park, Maryland, USA. Email: \tt\small\{mftariq,nileshs,mavridis,baras\}@umd.edu.}
\thanks{Research partially supported by ONR grant 
N00014-17-1-2622.}
\thanks{$^*$Faizan M. Tariq and Nilesh Suriyarachchi  are  co-first  authors.}
}

\begin{document}

\maketitle
\thispagestyle{empty}
\pagestyle{empty}
\begin{abstract}
Safe overtaking, especially in a bidirectional mixed-traffic setting, remains a key challenge for Connected Autonomous Vehicles (CAVs). The presence of human-driven vehicles (HDVs), behavior unpredictability, and blind spots resulting from sensor occlusion make this a challenging control problem. To overcome these difficulties, we propose a cooperative communication-based approach that utilizes the information shared between CAVs to reduce the effects of sensor occlusion while benefiting from the local velocity prediction based on past tracking data. Our control framework aims to perform overtaking maneuvers with the objective of maximizing velocity while prioritizing safety and passenger comfort. Our method is also capable of reactively adjusting its plan to dynamic changes in the environment. The performance of the proposed approach is verified using realistic traffic simulations.

\end{abstract}

\section{INTRODUCTION}

The promise of increased safety, efficiency and ease of access are the key motivations in the development and introduction of connected autonomous vehicles (CAVs) into modern road networks. While the situation where all vehicles on the road are fully autonomous remains a long term goal, it is likely that most initial CAVs introduced will need to operate side by side with human driven vehicles (HDVs) resulting in a mixed traffic situation. This results in many additional challenges brought about by the lack of cooperation and  unpredictability of human drivers \cite{mixedTraffic}. Overtaking on the incoming lane is a scenario where these issues play a significant role due to the increased possibility of head on collisions. 

This scenario is further complicated by low visibility, caused by sensor occlusion - a situation in which neighboring vehicles block the line-of-sight view of vehicles ahead and in other lanes. Due to this, there are several blind spots in an autonomous vehicle's perception of its environment. This then leads to either overly conservative behavior which reduces efficiency or highly risky maneuvers which may lead to increased collisions. The solution to this problem involves finding ways in which to bridge these gaps and fill in the missing information in a CAVs field of vision. To this effect, we explore the possibility of using vehicle to vehicle (V2V) communication between CAVs to share local information about neighboring tracked vehicles. The on-board sensor suite on CAVs is capable of detecting and tracking the dynamics of their immediate neighboring vehicles. This information can then be shared with other CAVs within the communication range. This means that in addition to its own on-board sensors, CAVs can collect information about other on-road vehicles and obstacles by communicating with CAVs downstream of its location.

\subsection*{Literature review}

The task of overtaking for a single agent has been studied extensively in literature with diverse control approaches. Sampling-based methods such as \cite{rrtFrazzoli} and \cite{rrt} provide safety guarantees, but can only provide asymptotic guarantees on the discovery of a suitable trajectory. Optimization-based approaches \cite{mpcDriving} and \cite{optimalControl} yield good performance but are often computationally expensive depending on the complexity of the models used for the vehicle dynamics. Extensions of these also explore the use of Robust MPC \cite{dixitRobustMPC} and Stochastic MPC \cite{stochasticMPC} in order to incorporate more realistic dynamic models and handle uncertainty in sensing and actuation. Learning-based methods \cite{drl} and \cite{mpcBayesCountry} have also been proposed for overtaking trajectory generation with real time operation capability but often lack safety guarantees. While these methods have not been applied to incoming lane overtaking, our previous work \cite{Faizan_overtake} explored the use of a mixed-integer model predictive control (MI-MPC) strategy for bidirectional overtaking for a single autonomous agent.

The use of communication among CAVs in order to improve the overall efficiency and safety of many complex traffic conditions such as highway merging \cite{Nilesh_merge} and traffic shock wave dissipation \cite{Nilesh_shockwave} have been studied in literature. This cooperation focused approach has multiple benefits in a bidirectional overtaking scenario. Some of these benefits have been studied in \cite{coop_risk} and \cite{coop_v2x} where probabilistic driven approaches have been applied to cooperative single direction overtaking and lane changing respectively.

\subsection*{Contribution}

The proposed method in this paper aims to combine the benefits of cooperation driven communication-based CAV control systems with the safety guarantees of a MI-MPC optimization-based controller for the domain of bidirectional overtaking in mixed traffic.

The main contribution of this work involves the implementation of a V2V communication-based multi-agent strategy for autonomous bidirectional overtaking which reduces the impact of blind spots created by sensor occlusion. In this regard, we also propose a velocity tracking method for trajectory prediction and an improved sensor occlusion model. 

Our method was tested using the realistic SUMO traffic simulation system, and the results and comparisons are presented. We show that our method leads to improved performance with less risky maneuvers, higher overall throughput and a high success rate for overtaking attempts. Further tests are also carried out to show the impact of CAV penetration levels and lane density on the performance of our method.

\section{Problem Description}

In the bidirectional overtaking problem, the controlled (ego) vehicle attempts to travel at its maximum safe velocity while safely overtaking the leading vehicles, if needed. In this section, we model the bidirectional overtaking scenario, road infrastructure used, characteristics of CAVs and HDVs, and the proposed sensor occlusion model. 

\subsection*{Notation} 

Throughout the manuscript, $\mathbb{R}$ denotes the set of real numbers. For some $a,c \in \mathbb{R}$ and $a<c$, we will write $\mathbb{R}_{[a,c]} = \{b \in \mathbb{R} \mid a \leq b \leq c \}$. 
Furthermore, $\mathbb{V}(k)$ will denote the set of all the vehicles in the simulation at time $k$ and $\mathbb{A}(k)$ will denote the set of all the CAVs in the simulation at time $k$. 

\subsection{Bidirectional overtaking scenario description}

\begin{figure} [h]
\centering
\includegraphics[trim=0 0 0 0, clip,width=.485\textwidth]{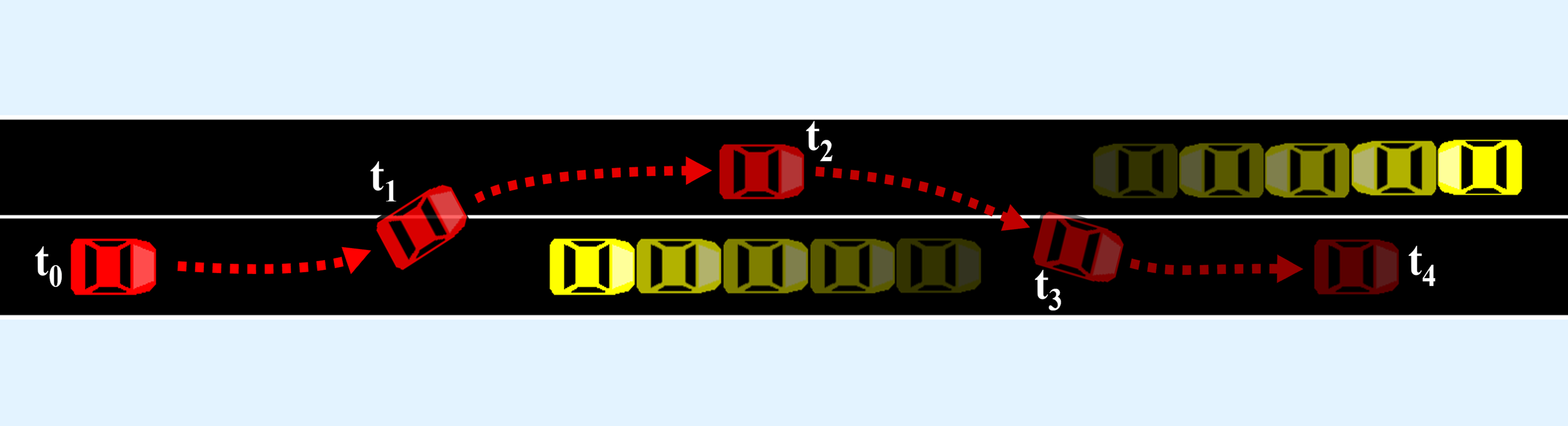}
\caption{Bidirectional traffic overtaking scenario.}
\label{fig:scenario}
\end{figure}

In the bidirectional overtaking scenario, an ego vehicle is required to overtake the leading vehicle(s) by moving into the adjacent lane while avoiding incoming traffic. This scenario involves three different types of vehicles: Autonomous ego vehicle which carries out the overtaking maneuver, the vehicles traveling ahead in the same lane as the ego vehicle and the vehicles approaching in the incoming lane. The non-ego vehicles can be either HDVs or CAVs. A simple three vehicle overtaking scenario with a sample trajectory is shown in Fig. \ref{fig:scenario}, with the ego vehicle depicted in red and the lead vehicle and incoming vehicle both depicted in yellow.

\subsection{Modeling the physical road structure}

\begin{figure} [ht] 
\centering
\vspace{0.6em}
\includegraphics[scale=0.178]{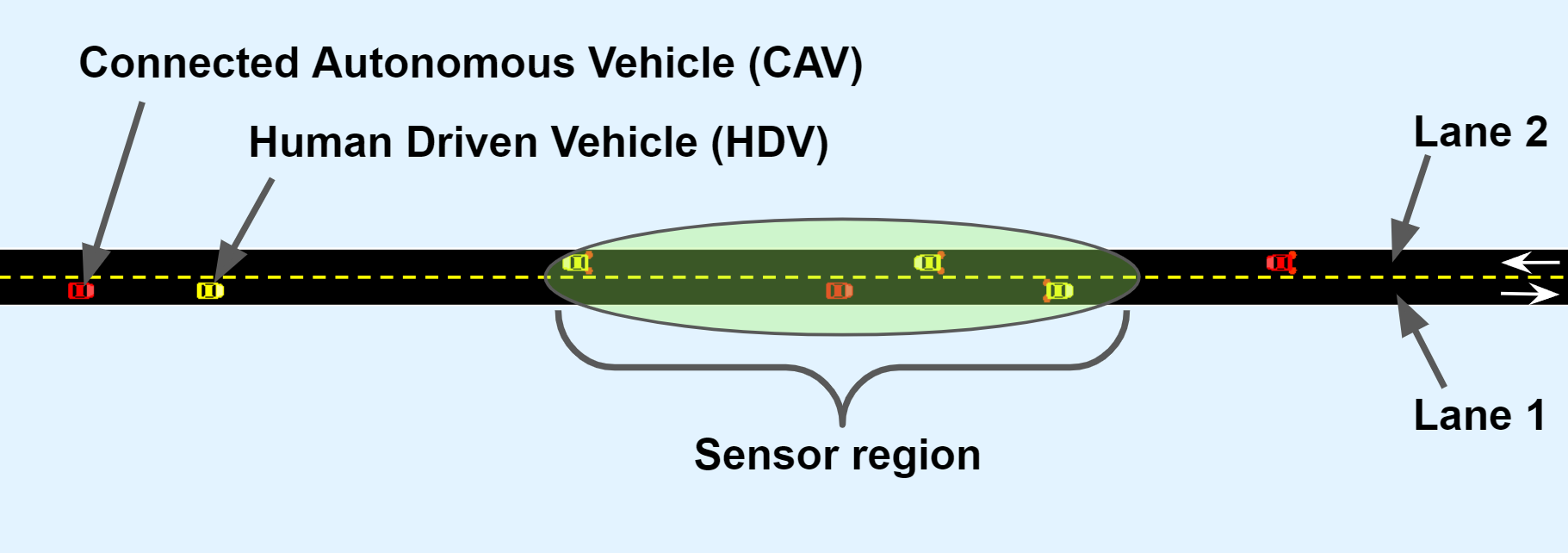}
\caption{Modeling a road section including vehicles.}
\label{fig: model}
\end{figure}

The physical road structure is modeled as a long continuous road segment with two adjacent lanes, each with traffic flow in opposite directions. We set the length of this road to $2$ km, which is a parameter that can be modified. We denote the speed limit of the road segment as $\bar v$ and compute distances in the Frenet coordinate system; the distance along the road is defined as the longitudinal displacement and the distance perpendicular to the road is defined as lateral displacement. 
It is also possible to change the density of vehicles on this road segment in either direction as necessary. These vehicles take the form of either CAVs or HDVs, and we can change the ratio of CAVs to HDVs (CAV penetration level) in the simulation. A section of this simulated road is shown in Fig. \ref{fig: model}, in which lane structure, HDVs, CAVs and CAV sensor ranges are highlighted. 

\subsection{Modeling CAV dynamics and control}
\label{sec: vehicle_model}


Two different types of vehicles are used in this research. Human driven vehicles whose motion is modeled according to section \ref{sec: traffic_model}, and CAVs that are modeled as follows.

In our research, we assume that each CAV has on-board, a low-level local controller $c_i$, which is capable of computing the necessary throttle and braking actuation commands in order to execute high level velocity goals as well as compute the steering commands that control the vehicles' lateral motion in order to keep the vehicle in lane. Therefore, for the proposed high level controller, we find it unnecessary to consider the highly non-linear dynamics of real-world vehicles.
This allows the $i^{th}$ vehicle to be modeled as a point object moving along the center of the lane according to the non-linear differential equation:
\begin{equation}
\label{eq:eq1}
        \dot{s_i} = f(t,s_i,c_i) , \qquad s_i(t^{0}_i) = s^{0}_i
\end{equation}
where $t_i^0$ is the initial time at which the $i^{th}$ vehicle enters the road segment.
Therefore, we can define the high-level discretized and linearized longitudinal vehicle dynamics by the following velocity control scheme:
\begin{equation}
\begin{aligned}
s_i(k) &= s_i(k-1) + \frac{v_i(k-1) + v_i(k)}{2} \cdot T_s \\
v_i(k) &= u_i(k) 
\end{aligned}
\label{eq:eq2}
\end{equation}
where $T_s$ denotes the sampling time while $s_i(k)$, $v_i(k)$, and $u_i(k)$ respectively denote the longitudinal displacement, velocity and applied control of each vehicle $i$ for $i \in \mathbb{A}(k) = \{1,\ldots,n(k)\}$. Here, $\mathbb{A}(k)$ represents the set of CAVs on the modeled highway stretch at time instant $k$ and the total number of CAVs is denoted by $n(k)$. It is important to note that the velocity control requested by the high-level controller should be reachable by the low-level vehicle controller $c_i$ in system (\ref{eq:eq1}). Additionally, as the control applied by the high level controller is independent of the lane the CAV is in, we also assume that the lane changing procedures are handled by a separate lane change controller. The input to the lane changing controller would be a high level goal indicating which lane to change into and what time to begin the lane change. For the overtaking case, the desired maneuver can be classified as two sequential lane changes at calculated times.

At a time instant $k$, each CAV $i$, is assigned an integer variable $l_i (k) \in\{0, 1\}$ which denotes the lane it is currently on (here, 0 and 1 represent the original lane and incoming lane respectively). It also has a vehicle length $L^e_i$ parameter, and has bounds on its maximum linear acceleration $A_i^{max}$ and maximum deceleration $A_i^{min}$ capabilities.
Therefore, the CAVs are defined by the following state vectors: 
\begin{equation}
\label{eq:eq3}
    X_i(k) = [s_i(k), v_i(k), l_i(k), L^e_i, A^{max}_i, A_i^{min}]^\T
\end{equation}

\begin{remark}
    For CAV $i \in \mathbb{A}(k)$, we define the sets $\mathbb{V}_0^i(k)$ and $\mathbb{V}_1^i(k)$, as the set of all the vehicles in the CAV's original lane ($l_i(k)$=$0$) and CAV's incoming lane ($l_i(k)$=$1$) respectively.
\end{remark}

\subsection{Modeling CAV sensing capabilities }
\label{sec: CAV_Sensing}

\begin{figure} [ht] 
\centering
\includegraphics[scale=0.17]{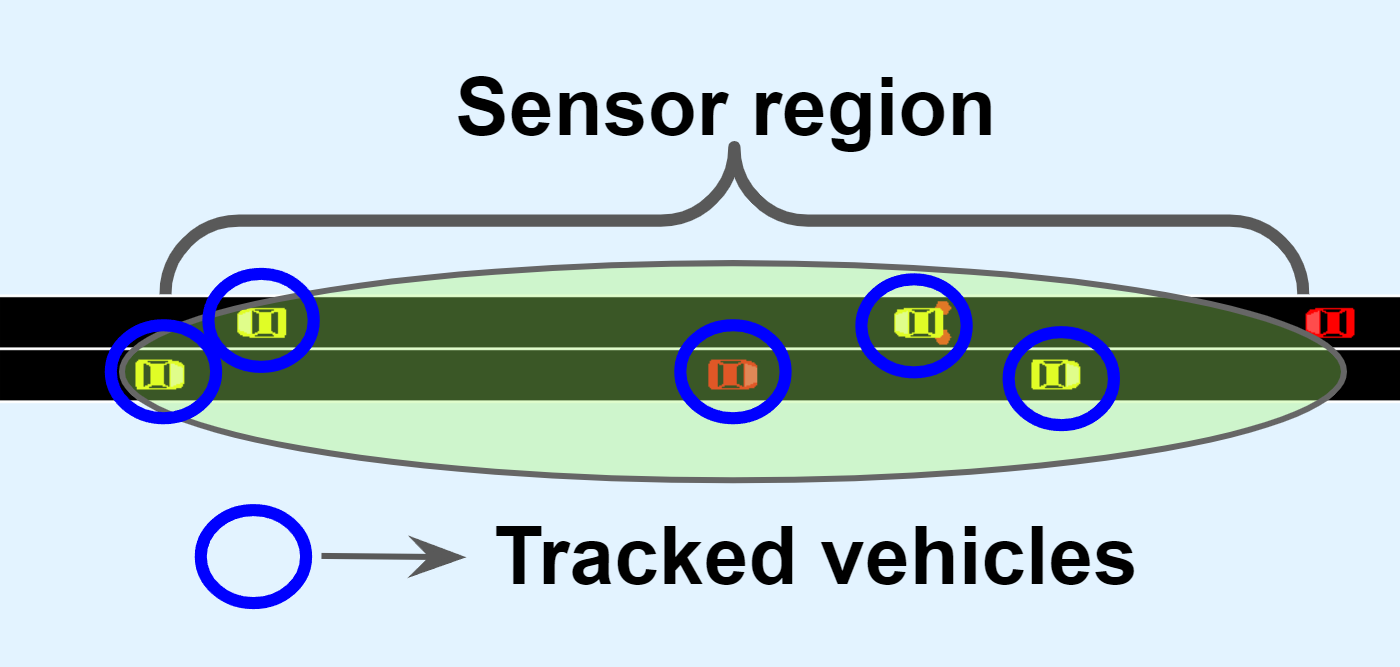}
\caption{Modeling sensing capability of CAVs.}
\label{fig: sensor_model}
\end{figure}

With regards to the sensing capabilities of the CAVs, each CAV is assumed to have the minimum required on-board sensing capability to detect the positions and velocities of surrounding vehicles within a realistic sensor range. Each CAV is assumed to be capable of tracking the positions of up to five adjacent non-occluded vehicles surrounding it. These vehicles involve the ego vehicles leader and follower in its own lane as well as a maximum of three vehicles in its adjacent incoming lane. In practice the actual number of vehicles tracked may be lower due to sensor occlusion and the density of vehicles on the road. A scenario where the ego vehicle is tracking four surrounding vehicles is highlighted in Fig. \ref{fig: sensor_model}.

This assumption of tracking surrounding vehicles is justified by the fact that modern CAVs have an advanced sensor suite which allows them to track the relative displacement of nearby line of sight obstacle/vehicles with very high accuracy. This data can then be used to compute the instantaneous velocities of surrounding vehicles with low margin of error. The main features of neighboring vehicles collected by each CAV are the vehicles longitudinal position, current lane, current velocity and a memory of past velocities. 


In addition to the on-board sensor suite, the other important source of information available to the ego vehicle is obtained in the form of V2V communication with other CAVs. Once each CAV uses its on-board sensors to collect the features of neighboring vehicles, it can then communicate this information with other CAVs within its communication range. Therefore, even if a vehicle is occluded to the ego vehicle, as long as this vehicle is visible to a downstream CAV, the ego vehicle can collect the information it needs for safe and effective trajectory planning.
When considering the V2V communication capabilities of each vehicle, we assume that the CAVs communicate using a combination of IEEE 802.11p and 5G networks. Additionally, in this research we do not consider the impact of network delay and packet loss during transmission. Therefore, we assume that vehicles within a realistic communication range of each other can share information in real time.  

\subsection{Improved sensor occlusion model}
\label{sec: sensorOcclusion}
\begin{figure} [h]
\centering
\includegraphics[trim=0 375 0 0,clip,width=0.475\textwidth]{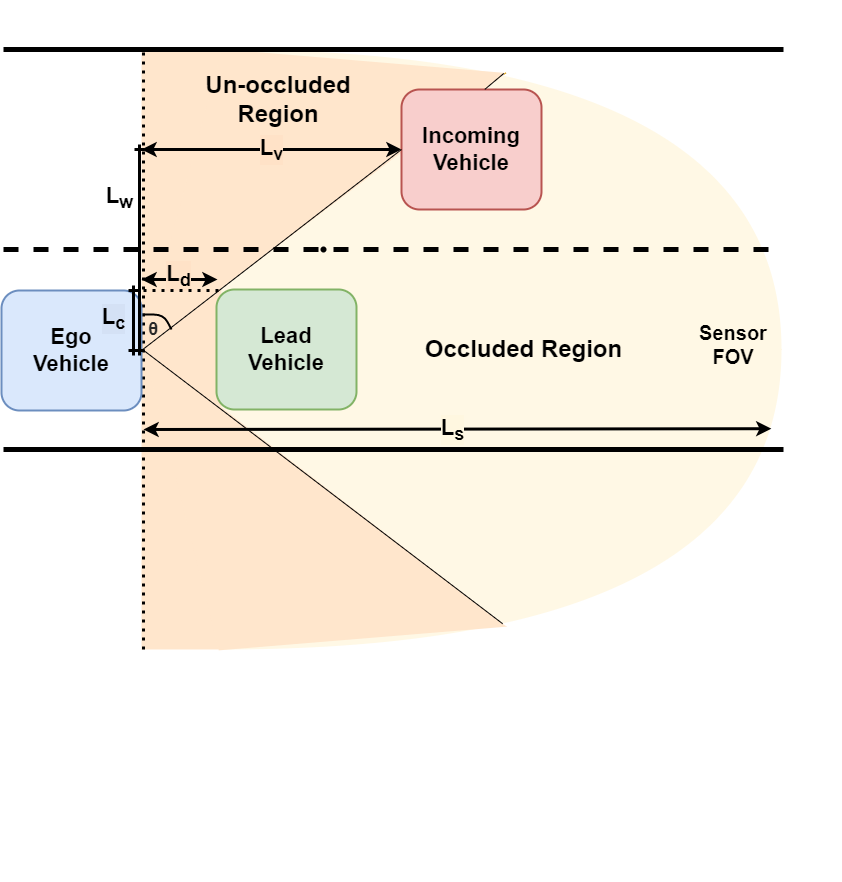}
\caption{Sensor Occlusion Model.}
\label{fig:occlusion}
\end{figure}

With the sensor occlusion model formulation, we aim to capture the effects of neighboring vehicles on the visibility range of a CAV. If there are no neighboring/blocking vehicles close to the ego vehicle, the visibility region is assumed to extend up to a fixed maximum value $L_s$, defined as the sensor range. However, not all vehicles within this range are visible to the CAV due to occlusion. The vehicle immediately ahead of the CAV will block the vehicles further up ahead. Additionally, depending on the proximity of the lead vehicle to the ego vehicle, visibility will also be reduced in the incoming lane. The resultant visible regions are shown in Fig. \ref{fig:occlusion}.

Here, $L_d^i(k)$, $L_c$, $L_w$ and $L_v^i(k)$ represent the distance gap to the leading vehicle, half the average width of a vehicle, the lane width, and the resulting un-occluded visible range in the adjacent lane respectively. These variables are connected as shown in (\ref{eq:occlusion_var}).
\begin{equation}
    z_p^i(k) = s_p(k) - s_i(k), \quad (p,i \in \mathbb{V}(k); ~ p \neq i)
\end{equation}
\begin{equation} \label{eqn:occlusion}
    L_d^i(k) = \min_{p \in \mathbb{V}(k)} \{ \{z_p^i(k), L_s \}  \mid l_p(k) = l_i(k) \}
\end{equation}
\begin{equation}
\begin{aligned} \label{eq:occlusion_var}
tan\theta &= \frac{L_d^i(k)}{L_c} = \frac{L_v^i(k)}{L_w} \\
L_v^i(k) &= \frac{L_d^i(k) \cdot L_w}{L_c}
\end{aligned}
\end{equation}

Given that $L_c$ and $L_w$ do not change with time, we can compute the instantaneous adjacent lane visibility range as a function of the distance gap to the leading vehicle. Vehicles present in this visibility region can be tracked by the ego vehicle and are added to the observation state of the ego vehicle.
\begin{subequations}
\begin{equation} \label{eqn:occlusion1}
        \mathbb{O}_i^0(k) = \{ p \in \mathbb{V}(k)  \mid l_p(k)=l_i(k), 0 \leq z_p^i(k) \leq L_d^i(k)\}
\end{equation}
\begin{equation} \label{eqn:occlusion2}
            \mathbb{O}_i^1(k) = \{ p \in \mathbb{V}(k)  \mid l_p(k) \neq l_i(k), 0 \leq z_p^i(k) \leq L_v(k)\}
\end{equation}
\begin{equation} \label{eqn:occlusion}
\begin{aligned}
    \mathbb{O}_i(k) &= \mathbb{O}_i^0(k) \cup \mathbb{O}_i^1(k)
\end{aligned}
\end{equation}
\end{subequations}

For an ego CAV $i \in \mathbb{A}(k)$, the set of observed vehicles (\ref{eqn:occlusion}) at time instant $k$, is denoted by $\mathbb{O}_i(k)$. The CAV $i$ can observe lead vehicles traveling ahead in its same lane (\ref{eqn:occlusion1}) as long as the lead vehicle is within the measurement range $L_s$. The CAV $i$ can observe vehicles traveling in its adjacent lane (\ref{eqn:occlusion2}), within the un-occluded visible region, $L_v(k)$ (\ref{eq:occlusion_var}).
 
\subsection{Modeling HDVs - Microscopic traffic models}
\label{sec: traffic_model}
The process of modeling human driving behavior in simulation usually involves two separate models. A car following model, used to compute the safe following velocity of a vehicle, considering its dynamic constraints and its interactions with the lead vehicle. A lane change model, used to determine when to change lane, and the parameters necessary for a safe lane change maneuver. While there are many car-following models such as the Krauss model \cite{Krauss} and the Intelligent Driver Model (IDM) \cite{IDM}, we opt to use the Krauss model for its accuracy and simplicity. This model computes the safe following speed $v_s(t)$ by considering the impact of speed limits $\bar v$, vehicle acceleration capabilities $a_{max}$, the vehicle deceleration profile $b(v(t))$, distance gap $\Delta s(t)$ and speed $v_l(t)$ of lead vehicle, time step $\Delta t$ and driver reaction time $\tau_r$ as shown in equation (\ref{eq:krauss}). The final output speed provided to the vehicle will then have a zero mean Gaussian noise added to it to model the imperfections in human driving. For the lane change model we use the Erdmann \cite{lc2013} model, which allows for the tuning of each vehicle's lane changing behavior.
\begin{equation}
\label{eq:krauss}
    v_s(t) = \min \left \{\bar v, v(t) + a_{max} \Delta t, v_l(t) + \frac{\Delta s(t) - v_l(t)\tau_r}{\frac{v(t)}{b(v(t))}+\tau_r} \right \}
\end{equation}

\section{Methods and Procedures}
In this section, we describe the state estimation algorithm to estimate the states of the observed vehicles, the prediction model to generate the predicted future trajectories for those vehicles, and the optimal control algorithm that is responsible for the decision making aspects of each of the CAVs.

\subsection{Tracked vehicle state prediction}
\label{sec: velPred}


In the literature, it has been shown that an arctangent function is a good representation of the acceleration model of a car. Therefore, we utilize a piecewise linear approximation of the arctangent function, to model the velocity profile of an observed vehicle. In order to predict the future velocity profile of a vehicle $p$ at time step $k$, we perform linear regression with mean-squared error on the previously observed velocity data points to obtain the slope ($\bar m_p^k$) parameter. We utilize the estimated parameter to project the velocity into the future for a given number of steps, defined as $H_a$. For the remaining duration of the prediction horizon, we assume the velocity to remain constant. This is compactly represented in (\ref{eqn:velEstimation}).
\begin{equation} \label{eqn:velEstimation}
\begin{aligned}
\hat v_p^k(0) &= v_p(k) \\
\hat v_p^k(j) &= \begin{cases}
\min \{\bar v, \hat v_p^k(j-1) + \bar m_p^k \cdot j \}, & 0 < j \leq H_a \\
\hat v_p^k(j-1), & H_a < j \leq H_p
\end{cases}
\end{aligned}
\end{equation}
Here, $\hat v_p^k(j)$ corresponds to the predicted velocity $j$ time steps into the future starting at time instant $k$ of an observed vehicle $p \in \mathbb{O}_i(k)$, $H_a$ corresponds to the acceleration horizon and $H_p$ corresponds to the prediction horizon. 

Each CAV then estimates the relative longitudinal displacement $\hat s_p^k(j)$ of the observed vehicles $p \in \mathbb{O}_i(k)$, (with $\hat s_p^k(0) = s_p(k)$), using the computed predicted velocities $\hat v_p^k(j)$ as follows:
\begin{equation} \label{eqn:posEstimation}
   \hat s_p^k(j) = \hat s_p^k(j-1) + \frac{\hat v_p^k(j-1) + \hat v_p^k(j)}{2} \cdot T_s 
\end{equation}


\subsection{Surrounding vehicle state aggregation}
\label{sec: totalSurroundingState}

Based on the state prediction model described in \ref{sec: velPred}, we know that each CAV can track a group of vehicles in its `line of sight' as discussed in \ref{sec: CAV_Sensing}. Note that the vehicles that can be tracked by a CAV also depend on the sensor occlusion status of the CAV as shown in \ref{sec: sensorOcclusion}. This tracked vehicle information, including predicted velocities and positions, can then be shared with other CAVs within V2V communication range. Therefore, the ego CAV $i \in \mathbb{A}(k)$ obtains information about its surrounding vehicles from two sources, its own on-board sensor systems ($\mathbb{O}_i(k)$) and the information communicated to it from other CAVs ($\mathbb{C}_i(k)$).

However, it is important that we do not overload the optimization-based controller with unnecessary information, which may lead to increased computation times. We introduce a sorting step to select which surrounding vehicles are most important for the overtaking problem. These vehicles (set $\mathbb{I}_i(k)$) that are pertinent to the decision-making process of CAV $i$, can include up to three leading vehicles and one following vehicle on the ego vehicle's own lane as well as up to two leading vehicles and one following vehicle in the incoming lane. Therefore, the predicted future states of up to six vehicles (set $\mathbb{W}_i(k)$) can be provided to the MI-MPC overtaking controller. The next step is to populate $\mathbb{W}_i(k)$ based on the information availability of vehicles in set $\mathbb{I}_i(k)$.  

\begin{equation} 
\label{eq:fullInformationSet}
    \mathbb{W}_i(k) = \mathbb{I}_i(k) \cap (\mathbb{O}_i(k) \cup \mathbb{C}_i(k))
\end{equation}

The MI-MPC controller of CAV $i$ will then be provided the predicted states of the vehicles in $\mathbb{W}_i(k)$. 

\subsection{Optimal model predictive control formulation}
\label{sec: optimalControl}
The proposed optimal controller is responsible for computing the sequence of velocity and lane change commands which would allow the ego CAV to maximize its velocity while respecting system dynamics and safety constraints.
In this formulation we also assume that the longitudinal and lateral dynamics of the ego CAV are decoupled \cite{decoupledDynamics}, as justified by the lower road curvatures present in highway overtaking scenarios and the capability of low-level lane change controllers to perform maneuvers while respecting lateral dynamic constraints \cite{scenarioMPC}.
Note that this proposed optimal controller is present on every CAV $i \in \mathbb{A}(k)$ and they all perform their own computations independently.

\subsubsection{Objective function} 
\label{sec:mimpc}
%

%

%
%
%
%
%
%

Each CAV $i \in \mathbb{A}(k)$ has an optimal controller, formulated as a mixed-integer model-predictive optimal control problem (MI-MPC), that provides, at any time instant $k$, the control input $u_i(k+1)$, and the binary overtaking decision $\mathcal{D}_i(k+1)$.

The objective function is formulated as a maximization of the velocity of the ego CAV while minimizing the time spent in the incoming lane and minimizing abrupt changes in velocity. This is defined as follows:

\begin{align}
\begin{split}
\min_{\substack{u^k_i(1), \cdots, u^k_i(H); \\ \mathcal{D}^k_i(1), \cdots, \mathcal{D}^k_i(H)}} \quad & \sum_{j=1}^H [-\gamma_1 \cdot u^k_i(j) + \gamma_2 \cdot \mathcal{D}^k_i(j) \\
&\qquad + \gamma_3 \cdot (u_i^k(j)-u_i^k(j-1))^2]
\label{eq:obj}
\end{split}
\end{align}

Here, H denotes the planning horizon. The proposed optimization objective (\ref{eq:obj}) contains three trade-off parameters $\gamma_1$, $\gamma_2$ and $\gamma_3$, which handle the trade-off between maximizing velocity, minimizing time spent in the incoming lane and minimizing sudden changes in velocity. Increasing $\gamma_1$ leads to increased focus on velocity maximization which results in more aggressive overtaking behaviors and a less comfortable experience for passengers. Increasing $\gamma_2$ and $\gamma_3$ on the other hand, leads to a reduction in risky overtakes while improving the comfort of the passengers, at the cost of increased travel time. 

\subsubsection{Dynamic constraints}

The dynamic constraints ensure that the optimization controller generates reachable controls.
At time instant $k$, the initial values of position, velocity control and lane control are set as $s^k_i(0)=s_i(k)$, $u^k_i(0)=v_i(k)$ and $\mathcal{D}^k_i(0)=l_i(k)$ respectively. 
Next, the longitudinal command velocity $u^k_i(j)$, for all $j\in\{1,\ldots,H\}$, 
is bounded by the speed limit: 
\begin{equation}
\label{eq:eq_cons_1}
    0\leq u^k_i(j)\leq \bar v
\end{equation}
as well as the acceleration capabilities of each vehicle:
\begin{equation}
\label{eq:eq_cons_2}
    A^{min}_i \cdot T_s \leq u^k_i(j)-u^k_i(j-1) \leq A^{max}_i \cdot T_s
\end{equation}
where $T_s$, as referenced previously, is the time resolution for our computations. 
Regarding the lateral movement, we confine the time-dependent binary decision variable $\mathcal{D}^k_i(j)$ to $\{0,1\}$.
More specifically, $\mathcal{D}^k_i(j)=1$ corresponds to the decision to travel in the adjacent lane while $\mathcal{D}^k_i(j) = 0$ represents the decision to travel in the original lane. A difference in the current lane and the binary decision variable (i.e.  $\mathcal{D}^k_i(1) \neq l_i(k)$) will trigger a lane changing maneuver at time instant $k$. 
We find that this simplification leads to a significant reduction in computational complexity.

\subsubsection{Safety constraints}

We next introduce the constraints responsible for the prevention of rear-end and lateral collisions. The longitudinal positions ($s_i(k)$) of the CAV are computed with respect to the decoupled longitudinal dynamics model given in equation (\ref{eq:eq2}).

At any future time step $j$ starting from time instant $k$, the ego CAV needs to maintain a safe longitudinal distance to all the known vehicles traveling in its current lane. This requirement can be defined as:
\begin{equation} \label{eqn:constSafety1}
\begin{aligned}
    (1-\mathcal{D}^k_i(j)) \cdot (&|\hat{s}^k_p(j) - s^k_i(j)| - (L^e_p + M^i_{s_p}(j))) \geq 0, \\
    & ~ \forall p \in \{\mathbb{V}_0^i(k) \cap \mathbb{W}_i(k) \}
\end{aligned}
\end{equation}
\begin{equation} \label{eqn:constSafety2}
\begin{aligned}
   \mathcal{D}^k_i(j) \cdot (&|\hat{s}^k_q(j) - s^k_i(j)| - (L^e_q + M^i_{s_q}(j))) \geq 0, \\
    & \qquad ~ \forall q \in \{\mathbb{V}_1^i(k) \cap \mathbb{W}_i(k)\}
\end{aligned}
\end{equation}
where $\mathbb{V}_0^i(k)$ and $\mathbb{V}_1^i(k)$ correspond respectively to the original lane and incoming lane vehicle set for the CAV $i$. Here, equations (\ref{eqn:constSafety1}) and (\ref{eqn:constSafety2}) represent the collision prevention constraints in the original and the incoming lanes respectively. These constraints need to be checked for all time instances in the planning horizon ($\forall j \in \{1,\cdots,H\}$). We define $M^i_{s_p}(j)$, the longitudinal safety margin that the ego CAV $i \in \mathbb{A}(k)$ needs to maintain from the vehicle $p \in \mathbb{V}(k)$ at time instant $k$, as follows:
\begin{equation}
    \begin{aligned}
         M^i_{s_p}(j) &= M_{0_p} + \frac{M_{v_p}}{\bar v} \hat v^k_p(j) + \frac{M_{a_p}}{A_i^{max}} \cdot \frac{|\hat v^k_p(j) - \hat v^k_p(j-1)|}{T_s} \\
        & + \mathds{1}_{\mathbb{V}^i_1(k)}(p) \cdot \frac{M_{l_p}}{\bar v} \cdot (v^k_i(j)+\hat v^k_p(j))
    \end{aligned}
\end{equation}
The safety margin $M^i_{s_p}(j) \in \mathbb{R}_{> 0}$ for vehicle $p$ depends on the longitudinal velocity, longitudinal acceleration, and the relative longitudinal velocity if $p$ is in the incoming lane of CAV $i$ ($p \in \mathbb{V}^i_1(k)$). Here, $M_{{0}_p} \in \mathbb{R}_{> 0}$ represents the nominal safety gap that will be maintained regardless of vehicle $p$'s driving behavior. Additionally, $M_{v_p} \in \mathbb{R}_{> 0}$, $M_{a_p} \in \mathbb{R}_{> 0}$ and $M_{l_p} \in \mathbb{R}_{> 0}$ correspond respectively to the multiplicative factors associated with the velocity, acceleration, and lane of vehicle $p$. 

To simplify the optimization process, we pose the safety constraints in the standard linear form, using multiple applications of the big-M method \cite{bigm}.
The optimization constraint (\ref{eqn:constSafety1}) is then converted to (\ref{eqn:constSafety1_1}) and (\ref{eqn:constSafety1_2}), while (\ref{eqn:constSafety2}) is converted to (\ref{eqn:constSafety2_1}) and (\ref{eqn:constSafety2_2}). This converts the optimization problem into a standard mixed integer quadratic program \cite{miqp}, which is computationally efficient to solve.

\begin{align} 
    \label{eqn:constSafety1_1}(\hat{s}^k_p(j) - s^k_i(j) + N_0 \, \cdot \, & a(j) - (L^e_p + M^i_{s_p}(j))) \nonumber \\ 
    & ~ + N_1 \cdot \mathcal{D}^k_i(j) \geq 0 \\
    \label{eqn:constSafety1_2}-(\hat{s}^k_p(j) - s^k_i(j) - N_0 \, \cdot \, & (1-a(j)) + (L^e_p + M^i_{s_p}(j))) \nonumber\\ 
    & ~ + N_1 \cdot \mathcal{D}^k_i(j) \geq 0 \\
    \label{eqn:constSafety2_1}(\hat{s}^k_q(j) - s^k_i(j) + N_0 \, \cdot \, & b(j) - (L^e_q + M^i_{s_q}(j))) \nonumber \\ 
    & ~ + N_2 \cdot (1 - \mathcal{D}^k_i(j)) \geq 0 \\
    \label{eqn:constSafety2_2}-(\hat{s}^k_q(j) - s^k_i(j) - N_0 \, \cdot \, & (1-b(j)) + (L^e_q + M^i_{s_q}(j))) \nonumber \\ 
    & ~ + N_2 \cdot (1 - \mathcal{D}^k_i(j)) \geq 0
\end{align} 
Here, these constraints are repeated $\forall p \in (\mathbb{V}_0^i(k) \cap \mathbb{W}_i(k))$, $\forall q \in (\mathbb{V}_1^i(k) \cap \mathbb{W}_i(k))$ and $\forall \ j \in\{1,\ldots,H\}$. Also, $N_0, N_1, N_2 \gg 0$, and $a(j), b(j) \in \{0,1\}$.
The constants $N_1$ and $N_2$ allow for automatic satisfaction of the inactive constraints based on the value of $\mathcal{D}^k_i(j)$. The constant $N_0$, in conjunction with the boolean variables $a(j)$ and $b(j)$ allows for the transformation of the absolute relative distance constraints in (\ref{eqn:constSafety1}) and (\ref{eqn:constSafety2}) into linear constraints.

%
%
 
%
%

The constraints given by equations (\ref{eq:eq_cons_1}), (\ref{eq:eq_cons_2}), (\ref{eqn:constSafety1_1}), (\ref{eqn:constSafety1_2}), (\ref{eqn:constSafety2_1}) and (\ref{eqn:constSafety2_2}) along with the objective function in (\ref{eq:obj}), form the optimization problem for optimal CAV longitudinal command velocity and lane changing decision computation.
The output of the optimization problem at time instant $k$ is $\{u^{k*}_i(1), \cdots, u^{k*}_i(H), \mathcal{D}^{k*}_i(1), \cdots, \mathcal{D}^{k*}_i(H) \}$ which is applied in a receding horizon fashion, i.e., $u_i(k+1)=u^{k*}_i(1)$, and 
$\mathcal{D}_i(k+1)=\mathcal{D}^{k*}_i(1)$.

\begin{remark}
    The behavior of the algorithm can be modified by altering the safety margin ($M_{{0}_p}$, $M_{v_p}$, $M_{a_p}$ and $M_{l_p}$) parameters. In our tests, these parameters are tuned empirically to guarantee safe overtaking behavior whenever overtaking attempts are carried out. 
    Additionally, the proposed optimal controller does not explicitly account for the time taken by the low-level lateral controller to execute its desired maneuver.
    We note that this simplification does not introduce further safety concerns, given that $L_{{0}_i}$ has been tuned appropriately (e.g. chosen large enough), and that the ego CAV has the ability 
    to retract a lane-changing decision without having to move all the way to the center of the adjacent lane, which is achieved by the receding horizon framework of MPC.
\end{remark}


\subsubsection{Low-level lateral controller} \label{sec:lateral}

The proposed framework allows for the incorporation of any lane-changing model and low-level lateral controller found in the literature
\cite{laneChange, surveyLateralControl},
as long as this controller does not result in significant changes to the longitudinal dynamics of the system. 
An example of such a decoupled lateral controller is provided in \cite{stochasticMPC}.
%
%

\section{Experimental Setup and Results}
\label{sec:results}

\begin{figure} [ht] 
\centering
\includegraphics[scale=0.30]{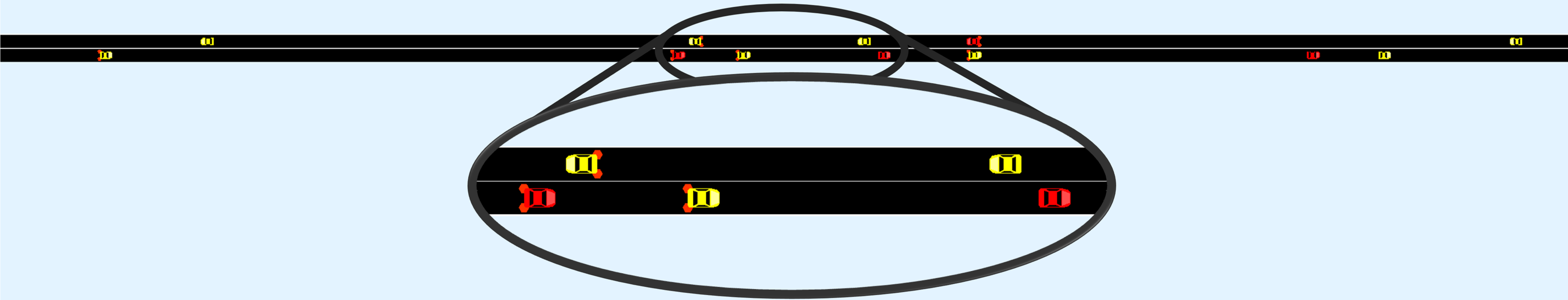}
\caption{Bidirectional road segment simulation.}
\label{fig: highway_sim}
\end{figure}

The performance of the proposed approach is evaluated on a bidirectional highway road segment simulation, implemented on the SUMO \cite{sumo} traffic simulation platform. 
The simulation setup for the highway segment is shown in Fig. \ref{fig: highway_sim}.
The controller communicates with the simulator using the TraCI traffic controller interface. 
All simulations and control algorithms are run on a personal computer with 
an Intel i7-8750H CPU and 32GB of RAM. 

The length of the bidirectional highway segment simulated is 2 km long. The vehicles used were a mix of CAVs and HDVs in varying proportions (CAV penetration levels). 

\begin{table}[h]
    \centering
    \begin{tabular}{|c|c|}
     \hline
     \textbf{Simulation Parameters} &  \textbf{Value} \\
     \hline
     Simulation step size & 100 ms \\
     Simulation duration & 1 hour \\
     Road length & 2 km \\
     Road speed limit ($\bar v$) & 20 m/s \\
     Average HDV speed & 10 m/s \\
     \hline 
     \textbf{Controller Parameters} &  \textbf{Value} \\
     \hline
     Controller sampling time ($T_s$) & 500 ms \\
     Maximum acceleration ($A^{max}$) & 4 $\text{m/s}^\text{2}$ \\
     Maximum deceleration ($A^{min}$) & -9 $\text{m/s}^\text{2}$ \\
     Maximum velocity ($V^{max}$) & 20 m/s \\
     CAV sensing range ($L_s$) & 150 m\\
     Planning horizon ($H$) & 10 s\\
     Safety Margin Parameters ($[L_{0_i}, L_{v_i}, L_{a_i}, L_{l_i}]$) & [10, 5, 5, 10]\\
     \hline 
    \end{tabular}
    \caption{Simulation \& Controller Parameters.}
    \label{tab:param}
    \vspace{-1.8em}
\end{table}

\subsection{Proposed method performance}

The performance of our proposed communication-based overtaking algorithm is evaluated according to its ability to avoid collisions, perform successful overtakes in order to maximize throughput and minimize the amount of failed overtake attempts and risky maneuvers. 
\begin{figure} [h]
\centering
\includegraphics[width=.48\textwidth]{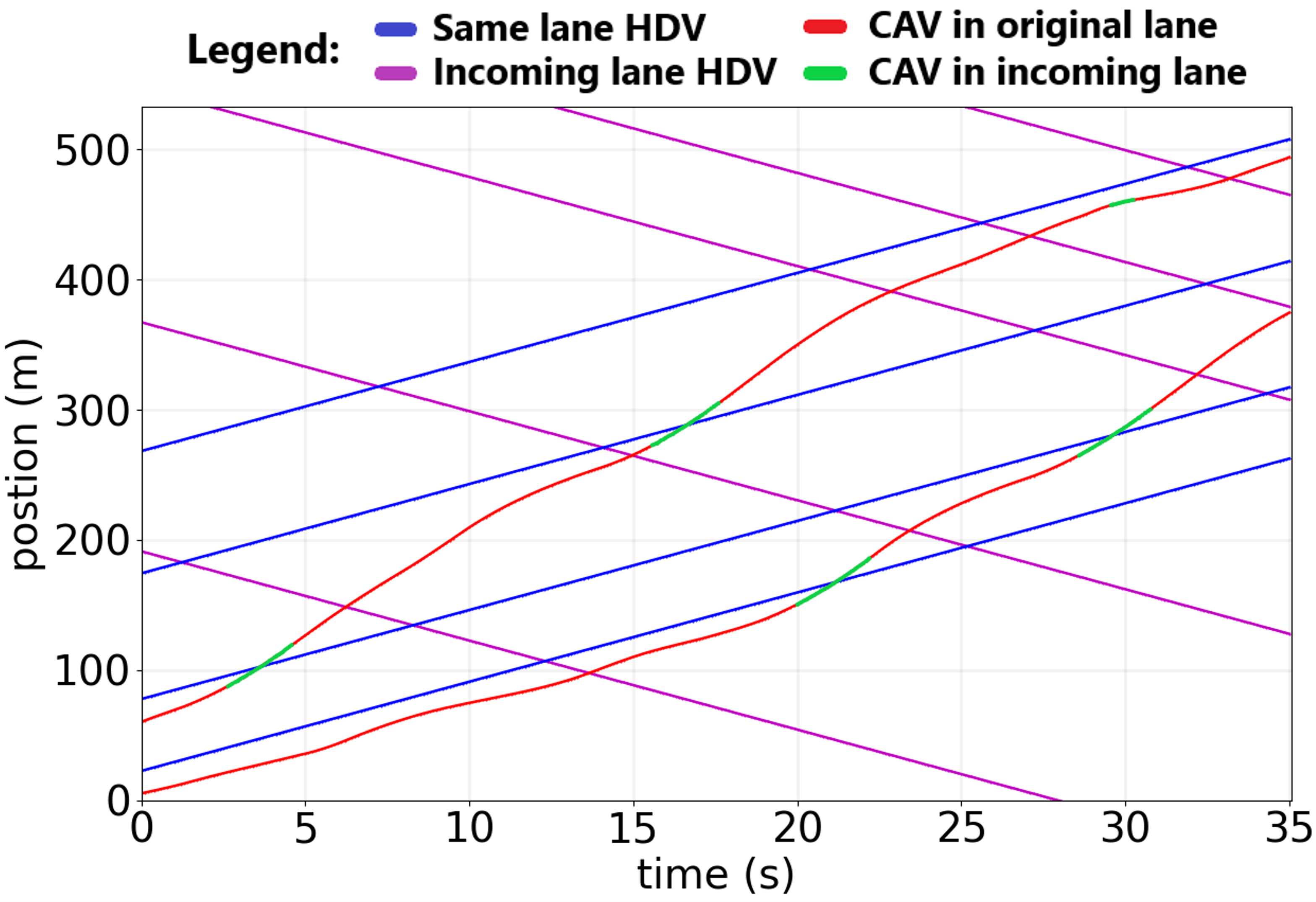}
\caption{Vehicle trajectories of $2$ CAVs attempting to overtake in the presence of $9$ HDVs. Overtakes occur when red CAV trajectory crosses blue HDV trajectory.
}
\label{fig:trajectories}
\end{figure}
Fig. \ref{fig:trajectories} shows the trajectories taken by 11 vehicles, 2 of which are CAVs following our control algorithm. The trajectories of these 2 CAVs are marked in red with the color switching to green whenever these CAVs move into the incoming lane. The blue trajectories having positive gradient depict the HDVs traveling in the same direction and lane as the CAVs. The purple trajectories having negative gradient depict the HDVs traveling in the opposite direction to the CAVs in the incoming lane. To avoid confusion we selected a section of the simulation in which no incoming lane CAVs were present. Overtakes occur whenever the CAV trajectory crosses a same lane HDV denoted in blue. Fig. \ref{fig:trajectories} shows four such successful overtakes. Note that at the point of overtaking, the trajectories of the CAVs should be green indicating that the CAV is in the incoming lane. As such collisions with CAVs occur only if a blue trajectory intersects a red trajectory (collision in original lane) or if a purple trajectory intersects a green trajectory (collision in incoming lane). We do not observe any of these conditions which shows us that our algorithm allows for collision free successful overtaking with the minimum time spent in the incoming lane. 

Additionally, we note that the average time required for each optimization computation step for an ego CAV is $32.36 \, ms$ with a standard deviation of $13.19 \, ms$. As the controller operates with a time step of $500ms$, this provides excess margins to ensure real time operation capability.

\subsection{Comparison between methods}
\begin{figure} [h]
\centering
\includegraphics[width=0.47\textwidth]{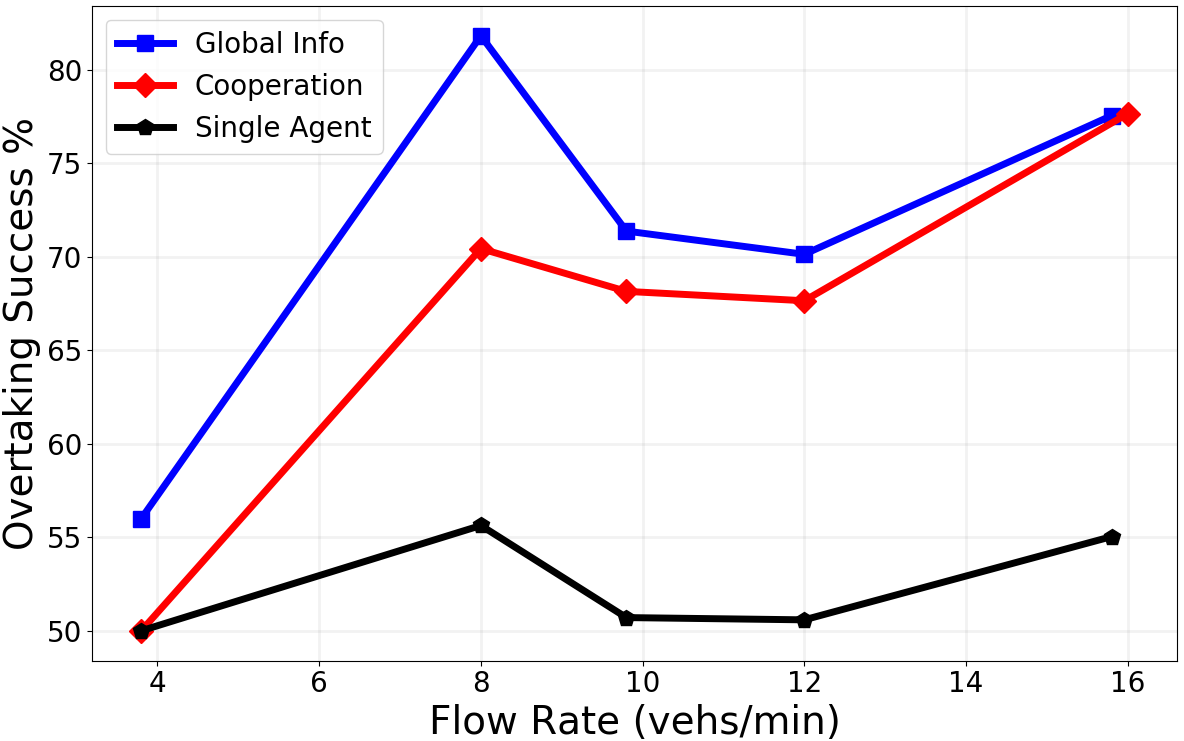}
\caption{Overtaking success over varying flow rate.}
\label{fig:flowSuccess}
\end{figure}

\begin{figure} [h]
\centering
\includegraphics[width=0.47\textwidth]{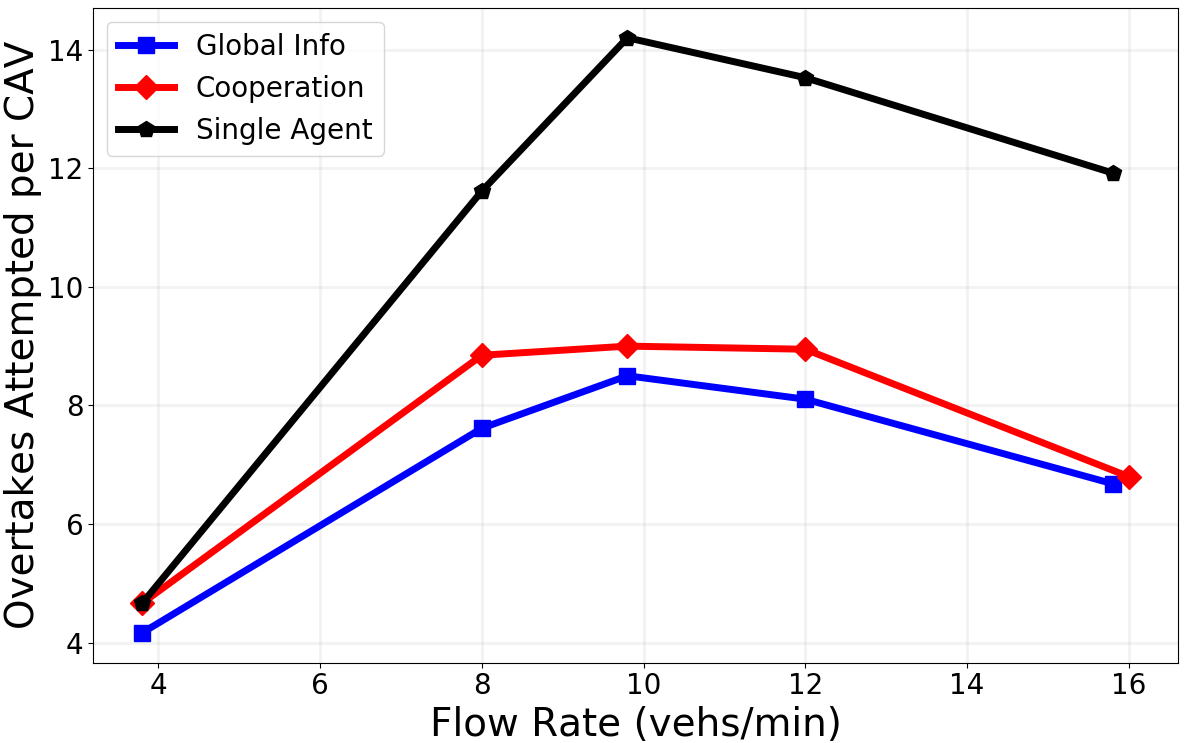}
\caption{Overtakes attempted over varying flow rate.}
\label{fig:flowAttempted}
\end{figure}

In order to highlight the benefits of the proposed communication-based approach, we compare its performance with an approach which does not use any inter-vehicle communication (Single Agent) and an approach which assumes the CAV has global knowledge (Global Info) about its surrounding. In the single agent approach, CAVs cannot communicate with each other and must rely on its own sensor information for decision making. In the global information case, CAVs make the unrealistic assumption of having access to the states of all its neighboring vehicles irrespective of sensor occlusion status. We utilize two key metrics to compare the performances of these three algorithms: Number of overtakes attempted per CAV and Success ratio of completed overtakes to attempted overtakes. Here, overtakes attempted tracks the number of times the CAV moves into the incoming lane to try to perform an overtaking maneuver. Overtaking success on the other hand tracks what percentage of these overtaking attempts actually lead to a successful overtaking maneuver. The performance of the three algorithms in regards to these two evaluation metrics with varying traffic flow conditions is depicted in Fig. \ref{fig:flowSuccess} and Fig. \ref{fig:flowAttempted}. Here, flow rate represents the vehicles per minute entering into the simulation. As the average velocities of vehicles  remains generally constant, flow rate is also directly proportional to the density of vehicles in the simulation. 

From Fig. \ref{fig:flowSuccess}, we observe that the cooperation based method as expected performs significantly better than the single agent method. This difference in performance is highlighted further with increasing levels of input flow rate. This is due to the fact that with increasing input flow rate, there is a higher probability that some of the vehicles within communication range of the ego CAV are other CAVs which can share information about their neighboring vehicles. This provides the ego CAV with more information about its surroundings and leads to less risky overtaking attempts and more overtaking successes. As expected the unrealistic global information method outperforms our cooperative method. However, we observe that this difference reduces with increasing input flow rate for the same reason involving increased number of CAVs in communication range. From Fig. \ref{fig:flowAttempted}, we observe that up to a certain input flow rate, all methods show an increase in overtake attempts since at higher densities there are more HDVs to overtake. However, beyond a certain level of flow rate we see a overall drop in overtaking attempts as there are less overtaking opportunities due to increased vehicle density. We also find that our method is able to gather enough information about its surroundings using communication, resulting in very low levels of risky overtakes attempted as witnessed by its comparative performance to single agent and global info strategies in Fig. \ref{fig:flowAttempted}.

\subsection{Impact of CAV penetration levels}

For our cooperative control strategy we find that the CAV penetration level plays a significant role in performance output. 

\begin{figure} [h]
\centering
\includegraphics[width=0.47\textwidth]{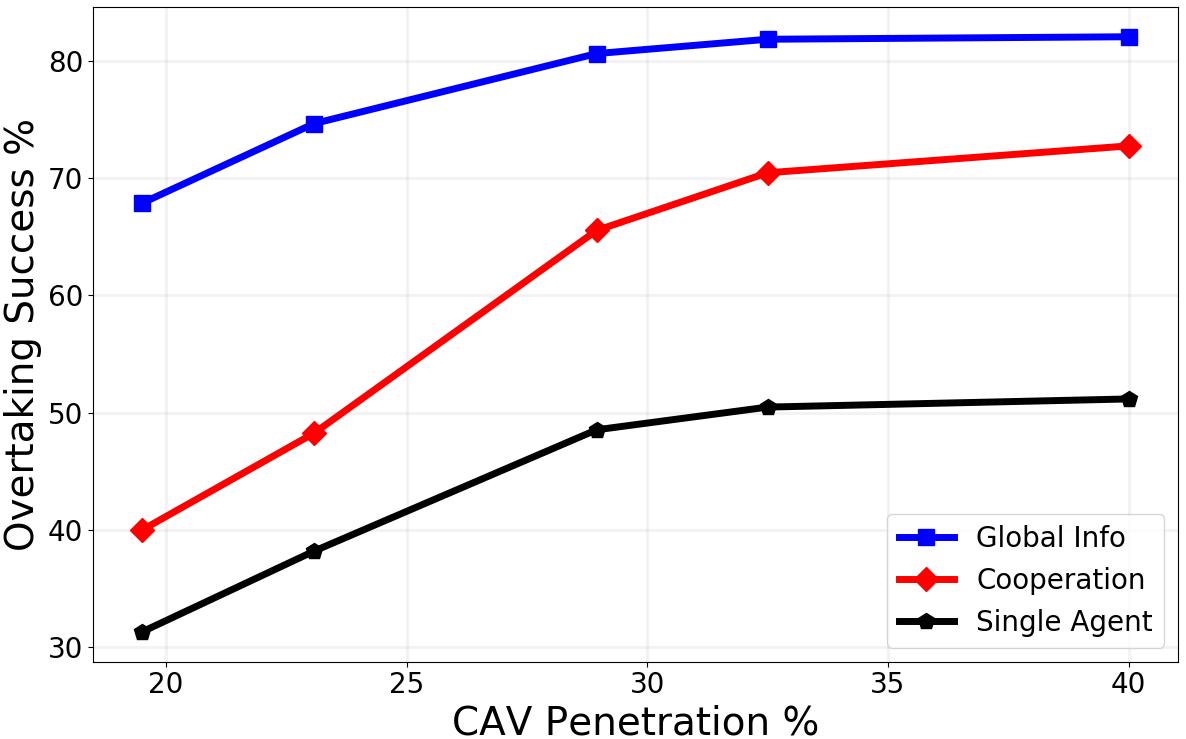}
\caption{Overtaking success over CAV penetration level.}
\label{fig:penetrationSuccess}
\end{figure}
\begin{figure} [h]
\centering
\includegraphics[width=0.47\textwidth]{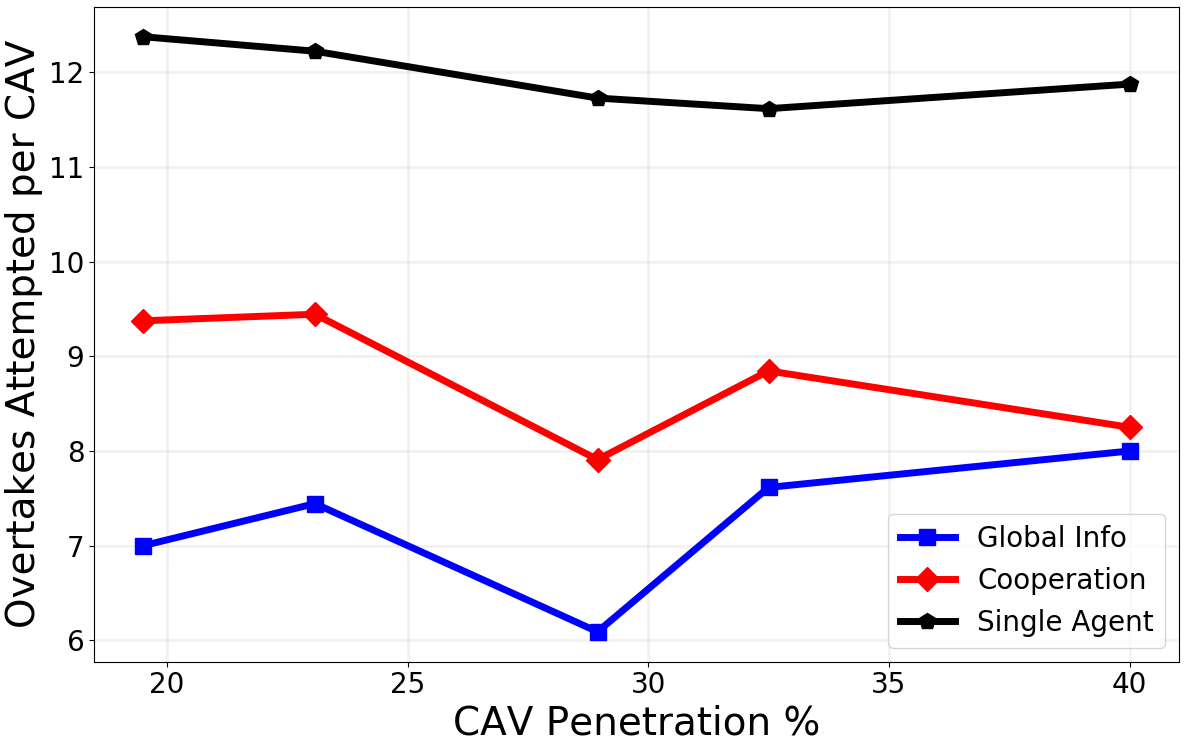}
\caption{Overtakes attempted over CAV penetration level.}
\label{fig:penetrationAttempted}
\end{figure}
Fig. \ref{fig:penetrationSuccess} shows an increase in successful overtakes in the cooperative method with increasing CAV penetration. This is due to the increase in the number of CAVs within communication range providing information to the ego CAV (i.e. increased CAV penetration leads to increased information available for decision making). 

In Fig. \ref{fig:penetrationSuccess} and \ref{fig:penetrationAttempted} we observe that the overall performance of our cooperative algorithm lies in between that of the global info and single agent methods. On average we achieve around $40\%$ improvement over the single agent method in terms of successful overtakes and reduced unnecessary overtaking attempts. We also find that as the CAV penetration increases our method approaches the performance of the global information approach. 
\vspace{1.0em}
\section{Conclusion}
\label{sec:conclusion}

We propose a novel bidirectional overtaking method for CAVs in mixed traffic, involving a V2V communication-based cooperative control strategy. In this method the CAVs share information with each other allowing them to overcome blind spots in sensing and perform safer overtaking maneuvers.
We couple the capabilities of V2V information sharing for traffic state estimation with a mixed-integer model predictive controller, capable of computing safe overtaking trajectories, in a bidirectional mixed-traffic setting.
We also perform explicit modeling of limited sensor ranges and blind spots caused by sensor occlusion to allow for realistic dense traffic performance tests of our approach. 
%
%
The performance of this method is evaluated using the SUMO platform and we demonstrate that this method is capable of achieving a much higher percentage of successful overtakes while reducing the amount of risky unnecessary overtaking attempts when compared to a single agent method with no communication between agents.
Future work could involve collaborative decision making among CAVs and exploration of learning-based approaches to control for bidirectional overtaking.


\bibliographystyle{IEEEtran}
\bibliography{References}

\end{document}